\documentclass{l4dc2026}
\jmlrproceedings{}{}
\jmlrvolume{}
\jmlryear{}
\usepackage{amssymb, amsfonts}
\usepackage{algorithm}
\usepackage[noend]{algorithmic}
\usepackage{geometry}
\usepackage{lipsum}
\usepackage{xcolor}
\usepackage{empheq}
\usepackage[most]{tcolorbox}
\usepackage{booktabs}
\usepackage{multicol}
\usepackage{multirow}
\usepackage{wrapfig}

\makeatletter
\@ifundefined{lemma}{}{ }
\@ifundefined{definition}{}{ }
\@ifundefined{assumption}{}{ }
\@ifundefined{remark}{}{ }
\makeatother

\newtheorem{lemma}{Lemma}                  
\newtheorem{definition}{Definition}       
\newtheorem{assumption}{Assumption}       
\newtheorem{remark}{Remark}               

\title[Hamilton-Jacobi Reachability and Conformal Prediction]{Statistically Assuring Safety of Control Systems using Ensembles of Safety Filters  
and Conformal Prediction} 
\author{%
 \Name{Ihab Tabbara} \Email{i.k.tabbara@wustl.edu}\\
 \addr Washington University in St. Louis
 \AND
 \Name{Yuxuan Yang} \Email{
y.yuxuan@wustl.edu}\\
 \addr Washington University in St. Louis
 \AND
  \Name{Hussein Sibai} \Email{
sibai@wustl.edu}\\
\addr Washington University in St. Louis
}

\date{}
\newcommand{\ihab}[1]{\textcolor{black}{#1}}

\begin{document}

   \pagestyle{empty} 
   \pagestyle{plain}
\maketitle

\begin{abstract}

Safety assurance is a fundamental requirement for deploying learning-enabled autonomous systems. Hamilton–Jacobi (HJ) reachability analysis is a fundamental method for formally verifying safety and generating safe controllers.
However, computing the HJ value function that characterizes the backward reachable set (BRS) of a set of user-defined failure states  is computationally expensive, especially for high-dimensional systems, motivating the use of reinforcement learning \ihab{approaches to approximate the value function}. Unfortunately, a  learned value function and its corresponding safe policy are not guaranteed to be correct.
 The learned value function evaluated at a given state may not be equal to the actual safety return achieved by following the learned safe policy. To address this challenge, we introduce a conformal prediction-based (CP) framework that bounds such uncertainty. 
 We leverage CP to provide probabilistic safety guarantees when using learned HJ value functions and policies to prevent control systems from reaching failure states. Specifically, we use CP to calibrate the switching between the unsafe nominal controller and the learned HJ-based safe policy and to derive safety guarantees under this switched policy. \ihab{We also investigate using an ensemble of independently trained HJ value functions as a safety filter and compare this ensemble approach to using individual value functions alone.}
 

\end{abstract}

\begin{keywords}%
  \ihab{Safety filter, Conformal prediction, Safe control, Ensemble learning}
\end{keywords}
\section{Introduction}

Verifying and synthesizing safe controllers for autonomous systems is essential, particularly as these systems increasingly operate in critical domains such as autonomous driving \citep{end_to_end_AD_survey_2024}, aerospace \citep{Runtime_assurance_for_safety-critical_systems}, and surgical robotics (\cite{autonomy_surgical_robots}). HJ reachability analysis has served as one of the tools for addressing this challenge. Given a failure set of states $\mathcal{F}$ that the system must avoid reaching, HJ reachability computes the corresponding backward reachable set (BRS), which is the set of all states from which reaching  the failure set cannot be prevented despite best control effort. Complementarily, it results in an 
optimal safe control policy that encodes the actions representing the best effort to avoid reaching the failure set.

Traditional methods for synthesizing HJ value functions based on dynamic programming or level set methods \citep{mitchell2005toolbox,stanfordaslhjreachability} scale exponentially with state dimension, becoming computationally challenging for high-dimensional systems. This curse of dimensionality has motivated researchers to use deep reinforcement learning (RL) approaches to approximate HJ value functions and their associated safe policies \citep{bansal2021deepreach,jaimes_claire_2019_hj_rl,bansal2017hamilton}. 
However, this scalability improvement comes at the cost of reliability: the learned HJ value function characterizing the backward reachable set 
may be inaccurate.
When evaluated at any state, its value might not be equal to the true return obtained by following the  policy that is trained to maximize it. 

To address this challenge, we use conformal prediction (CP), a statistical framework that provides distribution-free and finite-sample guarantees for predictions obtained from arbitrary black-box models \citep{angelopoulos2021gentle}. CP has been increasingly valuable in assuring the behaviors of robots and autonomous systems \citep{lindemann2024formalverification_conformal_prediction,lindemann2023safe_planning_usingconformalprediction}, with successful applications in providing  formal guarantees for safety filters with learned components~\citep{huriot2025safe,strawn2023conformal_safetyfilters_rl,somil2024reachability,somil2025reachability}. The key motivation 
is that conformal prediction can transform point predictions from black box models into prediction intervals without strong distributional assumptions.

In this work, we leverage CP to introduce a principled way to switch between a nominal (potentially unsafe) policy and a learning-based safety-maximizing one 
co-trained with a HJ value function and derive guarantees on the safety of the resulting switched policy, i.e., that it prevents the system from reaching failure states. Our framework operates in two stages.
In the first stage, rather than relying solely on the HJ value function's estimate of safety to decide when to switch between the nominal and safe policy, the switching decision is guided by calibrated confidence bounds that contain the actual safety return that the learned safe policy can achieve with high probability. In the second stage, we apply conformal prediction again to provide probabilistic guarantees that the resulting switched policy keeps the system safe, i.e.,  with high probability, the system will not enter the failure set at any time in the trajectory. Together, these two stages form a hierarchical safety-assurance framework that restores formal statistical safety guarantees for learned HJ value functions used as safety filters.

\ihab{In addition to the two-stage conformal prediction framework, we also investigate an ensemble approach where multiple HJ value functions are independently trained and calibrated using conformal prediction. We explore different methods for aggregating the safe policies from individual models and evaluate whether the ensemble-based safety filter achieves better safety guarantees than using individual models alone.}

\ihab{Our main contributions are: (1) We introduce a two-stage conformal prediction framework that provides probabilistic safety guarantees for switching between a nominal (potentially unsafe) controller and a learned HJ-based safe controller. (2) We propose, to the best of our knowledge, the first safety filter framework that integrates an ensemble of independently trained HJ value functions, investigating how to aggregate the safe policies from individual ensemble members.}

\section{Preliminaries}
\label{sec:preliminaries}
In this section, we provide the necessary background related to HJ reachability analysis and conformal prediction.
\subsection{Hamilton–Jacobi reachability analysis} 
HJ reachability is a control-theoretic approach  used for control synthesis and safety verification for general nonlinear control systems~\citep{HJ_Bansal_somil_claire_2017}. 
Consider a discrete-time nonlinear control system of the form $x_{t+1} = f(x_t, u_t)$, where $x_t \in X$ and $u_t \in U$. We denote the trajectory of the system starting from a state $x$ and following a policy $\pi: X \rightarrow U$ by $\xi_x^\pi: \mathbb{N} \rightarrow X$.  Given a set of states $\mathcal{F} = \{x \mid h(x) \leq 0\}$ that are considered {\em failure} states to {\em avoid}, where $h : X \to \mathbb{R}$, 
 HJ reachability analysis  computes the optimal 
 value function $V: X \to \mathbb{R}$, where $V(x) := \sup_{\pi \in \Pi}\inf_{t \in \mathbb{N}} h(\xi_x^\pi(t))$, which satisfies the fixed-point Bellman equation: $V(x) := \min \left\{ h(x), \max_{u \in U} V(f(x, u)) \right\}$.
 The associated safe policy derived from $V$ is $\pi_\text{safe}(x) := \arg \max_{u \in U} V(f(x, u))$. The backward reachable set (BRS) of the failure set $\mathcal{F}$ is defined as the set of all initial states from which the system is guaranteed to eventually reach $\mathcal{F}$ under any control sequence and is described as \{$x \in X \mid V(x) \leq 0$\}. It consists of the states starting from which the system will inevitably reach the failure set and are thus considered {\em unsafe}. 
 
 Consider a system that uses a nominal controller $\pi^{\text{nom}}$ that we wish to augment with an HJ reachability-based {\em safety filter}, i.e., a model that corrects any unsafe nominal actions. During deployment, the system follows $\pi^{\text{nom}}$ as long as $V(f(x, \pi^{\text{nom}}(x))) > 0$. Otherwise, the controller switches to $\pi_{\theta}^\text{safe}$, preventing the system from entering the BRS.

We use RL to approximate a time-discounted version of 
the HJ Q-function 
\citep{bridging_HJ_and_RL}.
Specifically, we use the DDPG \citep{ddpg} RL algorithm and optimize (\ref{eq:safety_q_loss}) to learn the Q-function and its corresponding safe policy. 

\begin{equation}
L(\theta) = {E}_{(x_t,u_t,x_{t+1})\sim D}\left[(Q_\theta(x_t, u_t) - y_t)^2\right]
\label{eq:safety_q_loss}
\end{equation}
where the target is computed as:
$
y_t = (1-\gamma)h(x_t) + \gamma \min\left\{h(x_t), \max_{u \in U} Q_{\theta}(x_{t+1}, u)\right\} 
$, and $D$ is a dataset of $(x_t,u_t,x_{t+1})$ triplets.
After the Q-function and the optimal safety-preserving policy $\pi_{\theta}^\text{safe}$ are learned, 
the corresponding HJ value function evaluated at state $x$ is
$
V_\theta(x) = Q_{\theta}(x, \pi_\text{safe}(x)). 
$

\subsection{Conformal prediction}
Conformal prediction is a distribution-free framework for uncertainty quantification that provides finite sample coverage guarantees for predictions from arbitrary black-box models, including machine learning-based ones (\cite{vovk2005algorithmic}). We provide a brief overview of {\em split conformal prediction}, a computationally efficient variant of the general conformal prediction method following (\cite{angelopoulos2021gentle}), and focusing on the case relevant to our application.

Given an input space $\mathcal{X}$ and output space $\mathcal{Y}$ of the function being approximated\footnote{The sets $\mathcal{X}$ and $\mathcal{Y}$ in this section are different from $X$ and $Y$ in the previous section, i.e., they are not necessarily the state space and the $Q$-learning targets.}, split conformal prediction assumes access to a calibration dataset $\mathcal{D}_{\text{cal}} = \{(x_i, y_i)\}_{i=1}^{n}$ consisting of pairs of inputs in $\mathcal{X}$ and correct outputs in $\mathcal{Y}$ that is held out from training. For each calibration point, we compute a \emph{nonconformity score} $s_i = s(x_i, y_i)$ measuring prediction error of the black-box model $\hat{f}:\mathcal{X} \rightarrow \mathcal{Y}$ when evaluated at $x_i$ and its output $\hat{f}(x_i)$ is compared with the ground-truth $y_i$. For one-sided lower bounds, we use $s(x, y) = \max(0,\hat{f}(x) - y)$.
Given a desired miscoverage rate $\alpha \in (0, 1)$, the quantile is defined as follows: $\hat{q}(\alpha) := \text{Quantile}(\{s_1, \ldots, s_n\}, \frac{\lceil (n+1)(1-\alpha) \rceil}{n})$. For a new test point $x_{\text{new}}$, the {\em one-sided prediction interval} is defined as follows: $c(x_{n+1},\alpha) := [\hat{f}(x_{n+1}) - \hat{q}(\alpha), \infty)$.

The guarantees of conformal prediction rely on the \textbf{exchangeability assumption}: the calibration set of pairs of random variables $(X_1, Y_1), \ldots, (X_n, Y_n)$ and test pair of random variables $(X_{n+1}, Y_{n+1})$, with sampling spaces $\mathcal{X}$ and $\mathcal{Y}$, must be exchangeable, i.e., for any permutation $\sigma$ of the indices $\{1, \dots, n+1\}$, $\mathbb{P}\big((X_1, Y_1), \dots, $ $ (X_{n+1}, Y_{n+1})\big) = \mathbb{P}\big((X_{\sigma(1)}, Y_{\sigma(1)}), \dots, (X_{\sigma(n+1)}, Y_{\sigma(n+1)})\big)$. 
Under this assumption, split conformal prediction provides the following \emph{coverage guarantee} 
\begin{equation}
\mathbb{P} \left( Y_{n+1} \in C(X_{n+1}, \alpha) \right) \geq 1 - \alpha
\end{equation}
where the probability is over the joint draw of the calibration set and test point and $C$ is the random variable representing the one-sided prediction interval resulting from the randomly sampled calibration set. 
This holds for any finite sample size $n$ and any model $\hat{f}$, without requiring any assumption on the data distribution besides exchangeability. 
For a fixed realized calibration set, 
the conditional coverage probability follows the Beta distribution Beta$(n + 1 - l, l)$, where $l = \lfloor (n+1)\alpha \rfloor$. Its mean is $\frac{n + 1 - l}{n + 2}$, which converges to $1 - \alpha$ as $n \to \infty$ (\cite{angelopoulos2021gentle}).

\section{Problem Setup}

We consider a discrete-time dynamical system governed by the state transition equation $x_{t+1} = f(x_t, u_t)$, where $x_t \in X \subseteq \mathbb{R}^n$ denotes the system state at time $t$, $u_t \in U \subseteq \mathbb{R}^m$ is the control input, and $f : X \times U \rightarrow X$ defines the system dynamics. The BRS of $\mathcal{F}$, denoted $X_{\text{unsafe}}$, is the set of unsafe states from which a safety violation is inevitable, i.e., despite best control effort, the system eventually reaches the failure set $\mathcal{F}$.  
The complement of this unsafe set defines the \emph{safe set} $X_{\text{safe}} = X \setminus (\mathcal{F} \cup X_{\text{unsafe}})$.
Our goal is to maintain the system inside $X_{\text{safe}}$ while primarily following a desired nominal controller $\pi^{\text{nom}}$, and to provide a probabilistic guarantee that the system will remain safe (i.e., the system will not enter $\mathcal{F}$).

\section{Methodology}
\label{sec:methodology}
We divide our framework into two phases. 

\vspace{0.07cm}
\noindent \textbf{Phase 1 (conformal prediction-based switching):} We use conformal prediction to calibrate the learned HJ value function and determine when to switch from the nominal controller $\pi^{\text{nom}}$ to the learned safe policy $\pi_{\theta}^\text{safe}$. Note that $\pi_{\theta}^\text{safe}$ is not guaranteed to be safety-preserving, but it represents the best policy at that task known to us during deployment. To achieve that, we define a one-sided nonconformity score that quantifies how much the learned HJ value function $V_\theta$ may overestimate safety, i.e., the actual value achieved by following $\pi_{\theta}^\text{safe}$, and construct a confidence interval around $V_{\theta}(x)$. We then define $\pi^\text{sw}$, where $\text{sw}$  refers to {\em switching}, as the data-driven policy that  
switches from $\pi^{\text{nom}}$ to $\pi_{\theta}^\text{safe}$ whenever the lower confidence bound on $V_{\theta}(f(x,\pi^{\text{nom}}))$ generated using conformal prediction is less than or equal to zero.

\vspace{0.07cm}
\noindent \textbf{Phase 2 (safety verification using conformal prediction):} Given the switching policy $\pi^\text{sw}$, we perform trajectory-level conformal calibration to certify its safety under the distribution of trajectories generated using $\pi^\text{sw}$. This stage provides a finite-sample, distribution-free guarantee that the system remains outside the failure set throughout the trajectory.

Together, these two phases form a framework that enables calibrated switching and provable statistical safety guarantees.

\subsection{Phase 1 (conformal prediction-based switching)}
First, we describe how the calibration dataset is collected and how the nonconformity scores are computed. We then present our switching algorithm. We first consider a single learned HJ value function and then consider an ensemble of separately trained and calibrated value functions.

\subsubsection{Calibration dataset construction}
We first collect a trajectory dataset $\mathcal{D}$ by executing the system under the switched policy that 
switches from $\pi^{\text{nom}}$ to $\pi_\theta^\text{safe}$ whenever the predicted next-state value satisfies $V_\theta(f(x_t, \pi^{\text{nom}}(x_t))) \leq  0$, and using $\pi^{\text{nom}}$ otherwise. From this dataset, we then construct a calibration set $\mathcal{D}_{\text{cal}}$ by sampling one state uniformly at random from each trajectory in $\mathcal{D}$ and omitting those trajectories from further consideration.
For each sampled state $x_i$, we compute $V_{\theta}^*(x_i)$, representing the actual safety value achieved by unrolling the safe policy $\pi_\theta^{\text{safe}}$ starting from $x_i$. In other words, we simulate the system forward using $\pi_\theta^{\text{safe}}$ and recursively calculate:
\[
V_{\theta}^*(x_i) = (1 - \gamma)\,h(x_i) + \gamma\,\min\{h(x_i),\,V_{\theta}^*(x_{i+1})\}.
\]

The recursion terminates when the trajectory either reaches the failure set $\mathcal{F}$, in which case $V_{\theta}^*(x_{i+k}) = h(x_{i+k})$, or when a user-defined horizon $H$ is reached, where we set $V_{\theta}^*(x_{i+H}) = V_{\theta}(x_{i+H})$.
For finite-horizon tasks, $H$ is set to the task's maximum horizon. For infinite-horizon problems, $H$ can be set large enough such that the discount factor renders future contributions beyond $H$ steps negligible. 

\subsubsection{One-sided calibration}
Since our safety decision is one-sided (checking if $V_{\theta}^*(x) > 0$), we define the nonconformity score as 
$
s(x_i) = \max(0, V_\theta(x_i) - V_{\theta}^*(x_i)).
$
When the model overestimates safety, i.e., $V_\theta(x_i) > V_{\theta}^*(x_i)$, the nonconformity score captures the magnitude of overestimation. When the model underestimates safety, i.e., $V_\theta(x_i) \leq V_{\theta}^*(x_i)$, the score is 0. With the nonconformity score, we obtain the lower bound of the conformal prediction interval.
\ihab{
\begin{definition}
\label{def:vlower}
For a given state $x$ and confidence level $1-\alpha$,  the lower bound of the conformal prediction interval is defined as
\[
\underline{V}(x, \alpha) = V_{\theta}(x) - \hat{q}(\alpha),
\]
where $\hat{q}(\alpha)$ is the $(1-\alpha)$-quantile of the nonconformity scores. 
\end{definition}
}


\subsubsection{Switched policy}
\noindent  Assuming that the states encountered at deployment are exchangeable with those in the calibration set, we present Lemma~\ref{lemma:single_step_coverage}. We acknowledge that this assumption may be violated due to distribution shift: the switching behavior between $\pi^{\text{nom}}$ and $\pi_{\theta}^\text{safe}$ during calibration dataset collection can differ from that induced by the deployed policy $\pi^\text{sw}$ (Algorithm~\ref{alg:single_model_safety_filter}) or $\pi^\text{sw}_{\text{ensemble}}$ (Algorithm \ref{alg:ensemble_safety_filter_strategies}) at test time, \ihab{where $\pi_{\text{sw}}$ is a policy that switches between $\pi^{\text{nom}}$ and the learned HJ-based safe controller $\pi_{\theta}^{\text{safe}}$ based on the lower bound of the conformal prediction interval, and $\pi_{\text{sw}}^{\text{ensemble}}$ applies the same principle using an ensemble of HJ value functions which will be discussed later in Section \ref{sec:ensemble_of_HJ}}.
 


\begin{algorithm}[H]
\caption{$\pi^\text{sw}$: Calibrated switching between $\pi_{\theta}^\text{safe}$ and  $\pi^{\text{nom}}$}
\label{alg:single_model_safety_filter}
\begin{algorithmic}[1]
\STATE \textbf{Inputs:} initial state $x_0$, nominal controller $\pi^{\text{nom}}$, 
safe controller $\pi_{\theta}^\text{safe}$, learned HJ value function $V$, 
calibrated quantile $\hat{q}(\alpha)$ 
\STATE $x \gets x_0$
\WHILE{termination condition is not met (e.g., horizon reached or task  goal achieved)}
    \STATE Predict next state under nominal control: 
    $x_{\text{next}} \gets f(x, \pi^{\text{nom}}(x))$
    \STATE Compute conformal lower bound: 
    $\underline{V}(x_{\text{next}}, \alpha) \gets V_{\theta}(x_{\text{next}}) - \hat{q}(\alpha)$
    \IF{$\underline{V}(x_{\text{next}}, \alpha) \ihab{>}  0$}
        \STATE $u \gets \pi^{\text{nom}}(x)$
    \ELSE
        \STATE $u \gets \pi_{\theta}^\text{safe}(x)$
    \ENDIF
    \STATE Wait until next time step and sense the new state $x$ 
\ENDWHILE
\end{algorithmic}
\end{algorithm}

\begin{lemma}
\label{lemma:single_step_coverage}
Given a user-defined miscoverage rate $\alpha \in (0,1)$, for any $t\in \mathbb{N}$,  
the prediction interval $\mathcal{C}(x_{t+1}, \alpha) := [V_{\theta}(x_{t+1}) - \hat{q}(\alpha), \infty)$ 
contains $V_{\theta}^*(x_{t+1})$ with probability at least $1 - \alpha$:
\[
\mathbb{P} \big(V_{\theta}^*(x_{t+1}) \in \mathcal{C}(x_{t+1}, \alpha)\big) \ge 1 - \alpha.
\]
This is a \textbf{marginal} guarantee: the probability is taken over the joint draw of the 
calibration set and the test point, conditional on the training split.

Conditioning on a calibration set $\mathcal{D}_{\text{cal}}$, 
the coverage for the next test point is not lower-bounded by $1-\alpha$. 
Following (\cite{angelopoulos2021gentle}), the conditional coverage random variable
\[
\Pr\!\big(V_{\theta}^*(x_{t+1}) \in \mathcal{C}(x_{t+1}, \alpha) \mid \mathcal{D}_{\text{cal}}\big)
\]
follows the Beta distribution $\mathrm{Beta}(N + 1 - l,\, l)$, where $l = \lfloor (N + 1)\alpha \rfloor.$
The mean of this distribution is $\tfrac{N + 1 - l}{N + 2} \approx 1 - \alpha$, 
indicating that for any fixed calibration set, the actual conditional coverage 
may lie above or below the nominal level $1-\alpha$, 
but concentrates around the target coverage as $N$ increases.
\end{lemma}

\subsubsection{Ensemble of HJ value functions}
\label{sec:ensemble_of_HJ}
Next, we construct an ensemble of independently trained HJ value functions, denoted $\{V_{\theta_j}\}_{j=1}^M$. We use the subscript $j$ to denote quantities and variables associated with the $j$-th ensemble member, such as its value function $V_{\theta_j}$, conformal quantile $\hat{q}_j(\alpha)$, and learned safe policy $\pi_{\theta_j}^{\text{safe}}$. 
For each member model $j$, the weights $\theta_j$ of $V_{\theta_j}$ are randomly and independently initialized. 
We then calibrate each model separately using conformal prediction to obtain its corresponding quantile $\hat{q}_j(\alpha)$. 
The goal of this ensemble setup is to examine whether combining multiple calibrated models can yield safer and less conservative performance than any single model alone or than their average.

To decide which member’s policy to follow at each step, we introduce two switching strategies. 
In the \texttt{single} switching strategy, whenever all ensemble members predict that the next state is unsafe under the nominal control, the agent switches to the safest member model 
defined as 
$j^{\ast} = \arg\max_{j \in \{1,\ldots,M\}} \underline{V}_j(x_{\text{next}}, \alpha)$,
and continues to use that model’s learned safe policy  $\pi_{\theta_{j^*}}^{\text{safe}}$ until the predicted next state when following the nominal controller is deemed safe again, i.e., $\underline{V}_{j^{\ast}}(x_{\text{next}}, \alpha) > 0$, where $x_{\text{next}} := f(x, \pi^{\text{nom}}(x))$.  
In contrast, in the \texttt{multiple} switching strategy, at each time step where the nominal controller is unsafe, the agent re-evaluates all ensemble members and selects the current safest model according to the same $\arg\max_{j}$ criterion, thereby allowing dynamic switching across ensemble members within a single unsafe episode.

The resulting ensemble safety filter $\pi^\text{sw}_{\text{ensemble}}$ is summarized in Algorithm~\ref{alg:ensemble_safety_filter_strategies}.

\begin{algorithm}[H]
\caption{$\pi^\text{sw}_{ensemble}$: calibrated switching between $\pi_{\theta_j}^{\text{safe}}$ and $\pi^{\text{nom}}$}
\label{alg:ensemble_safety_filter_strategies}
\begin{algorithmic}[1]
\STATE \textbf{Inputs:} $x_0$, $\pi^{\text{nom}}$, $\{V_{\theta_j}, \pi_{\theta_j}^{\text{safe}}, \hat q_j(\alpha)\}_{j=1}^M$, 
switching strategy $\in\{\texttt{single},\,\texttt{multiple}\}$
\STATE $x \gets x_0$, \quad $j_{\text{active}} \gets \bot$
\WHILE{termination condition is not met (e.g., horizon reached or task  goal achieved)}
    \STATE $x_{\text{next}} \gets f\big(x,\,\pi^{\text{nom}}(x)\big)$
    \STATE $\forall j$, $\underline{v}_j \gets V_{\theta_j}(x_{\text{next}}) - \hat q_j(\alpha)$
    \IF{$\max_j \underline{v}_j >  0$}
        \STATE $u \gets \pi^{\text{nom}}(x)$; \quad $j_{\text{active}} \gets \bot$
    \ELSE
        \IF{$\text{strategy} = \texttt{single}$ \text{and} $j_{\text{active}} \neq \bot$}
            \STATE $u \gets \pi_{\theta_{j_\text{active}}}^{\text{safe}}(x)$
        \ELSE
            \STATE $j_{\text{active}} \gets \arg\max_j \underline{v}_j$;
            \quad $u \gets \pi_{\theta_{j_\text{active}}}^{\text{safe}}(x)$
        \ENDIF
    \ENDIF
    \STATE Wait until next time step and sense the new state $x$
\ENDWHILE
\end{algorithmic}
\end{algorithm}

\subsection{Phase 2 (safety verification using conformal prediction)}\label{sec:phase2}

Phase 2 addresses the distribution shift problem by certifying the switching policy $\pi^\text{sw}$ and $\pi^\text{sw}_{ensemble}$ through trajectory-level analysis on the actual deployment distribution. Theorem \ref{thm:conformal_pi_star} formalizes the safety guarantee using $\pi^\text{sw}$ and $\pi^\text{sw}_{ensemble}$.

\begin{theorem}
\label{thm:conformal_pi_star}
Let $\pi^\text{sw}$ or $\pi^\text{sw}_{ensemble}$ be our switched policies from Phase 1 and $\pi \in \{\pi^\text{sw}, \pi^\text{sw}_{ensemble}\}$ and fix some $N_{\text{cert}} \in \mathbb{N}$. 
Consider $N_{\text{cert}}$ i.i.d. initial states $\{x_0^{(i)}\}_{i=1}^{N_{\text{cert}}}$ sampled from the deployment distribution $\mathcal{P}_0$ and corresponding trajectory safety margins under $\pi$ defined as follows:
\begin{equation}
J^{\pi}(x_0^{(i)}) = \min_{\tau \in [0,T_i]} h(\xi^{\pi}_{x_0^{(i)}}(\tau)),
\end{equation}
where $T_i$ is the termination time of the $i^{\mathit{th}}$ trajectory and  $\xi^{\pi}_{x_0^{(i)}}(\tau)$ is the state at time $\tau$ when starting from $x_0^{(i)}$ and following $\pi$.
Finally, let $k$ denote the number of trajectories with $J^{\pi}(x_0^{(i)}) \leq 0$ (the agent enters the failure set at some time instant). Then,
\[
\mathbb{P}_{x_0 \sim \mathcal{P}_0}\big(J^{\pi}(x_0) > 0\big) \sim \text{Beta}(N_{\text{cert}} - k, k + 1)
\]
\end{theorem}
The proof of the theorem is in Appendix~\ref{app:conformal_pi_star}.

\section{Case Study}
\label{sec:casestudy}
In this section, we examine our framework on a highway takeover environment. 
\subsection{Task description}
We use the 10-dimensional triple-vehicle highway takeover environment from (\cite{highway}), where the ego vehicle, modeled by nonlinear unicycle dynamics, is controlled to safely overtake the leading vehicle while avoiding the lateral vehicle driving in the opposite direction. Details related to the task,  the design of $\pi^\text{nom}$, and the definition of $\mathcal{F}$ in this task can be found in Appendix \ref{app:task}. A violation is defined as the ego vehicle colliding with another vehicle or leaving the road. A simulation is considered successful when the ego vehicle safely overtakes the two vehicles and reaches the end of the highway within $200$ time steps and without violating any constraints.  We train five HJ value functions $V_{\theta_j}$ and their corresponding $\pi_{\text{safe}}^j$ using DDPG. Each of the five models was trained separately using a different random seed for neural network parameter initialization, but with the same training hyperparameters. 
When calculating the nonconformity scores during the construction of $\mathcal{D}_\text{cal}$, where $|\mathcal{D}_\text{cal}|=10000$, we \ihab{consider a horizon-based termination condition ($H=200$) which is also used as the while loop termination condition  in Algorithms~\ref{alg:single_model_safety_filter} and \ref{alg:ensemble_safety_filter_strategies}. }

\subsection{Analysis of single-model safety filters}
\begin{figure}[h]
    \centering
    \includegraphics[width=1    \linewidth]{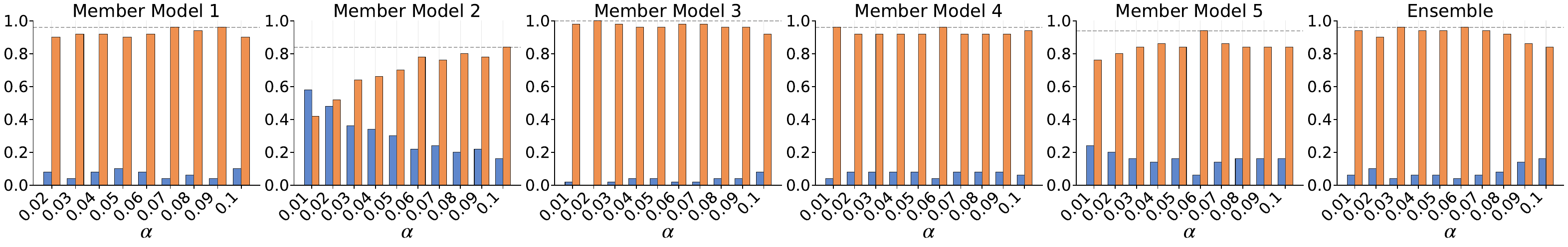}
    \vspace{-0.8cm}
    \caption{\small \textcolor{blue}{Violation rates} (blue bars) and \textcolor{orange}{success rates} (orange bars) achieved in 50 trials using an ensemble of HJ value functions as safety filters following the  \texttt{multiple} strategy in Algorithm \ref{alg:ensemble_safety_filter_strategies} and its corresponding  member models following Algorithm 
    \ref{alg:single_model_safety_filter} for different choices of  coverage rates $\alpha$.}
    \label{fig:robustness}
\vspace{-0.5cm}
\end{figure}

We first analyze the performance of each single-model HJ reachability-based safety filter. Observing Figure~\ref{fig:robustness}, most member models perform relatively well across different values of $\alpha$, with the notable exception of member model~2. Recall that smaller values of $\alpha$ produce more conservative switching. Hypothetically, this increased conservatism should lead to monotonically decreasing violation rates as $\alpha$ decreases. However, we do not observe this monotone trend in our experiments, likely due to two factors: (1) the limited sample size of 50 trials
and (2) distribution shift caused by using switching thresholds during deployment that differ from the threshold used when collecting data for  training the value functions.

\begin{table}[h]
\centering
\tiny
\begin{tabular}{cccccccccc}
\toprule
\multirow{2}{*}{member model} &  \multirow{2}{*}{1} &  \multirow{2}{*}{2} &  \multirow{2}{*}{3} &  \multirow{2}{*}{4} &  \multirow{2}{*}{5} & \multicolumn{2}{c}{Ensemble 1-5}&  \multirow{2}{*}{$\pi^\text{nom}$ only} \\
\cmidrule(lr){7-8}&&&&&&\texttt{multiple}&\texttt{single}\\
\midrule
Violation & 0.08 & 0.22 & 0.02 & 0.04 & 0.06 & 0.04 & 0.06 & 0.46\\
Success & 0.92 & 0.78 & 0.98 & 0.96 & 0.94 & 0.96 & 0.94 & 0.54\\
$\hat{q}(\alpha=0.06)$ & 1.61 & 2.37 & 0.87 & 1.04 & 1.16 & - & - & -\\
\bottomrule
\end{tabular}
\caption{\small Violation and success rates achieved across 50 trials by the nominal controller $\pi^\text{nom}$, each single HJ-based value function paired with $\pi^{\text{nom}}$ following Algorithm \ref{alg:single_model_safety_filter}, and the ensemble of HJ-based value functions paired with $\pi^{\text{nom}}$ following Algorithm \ref{alg:ensemble_safety_filter_strategies} for $\alpha=0.06$.}
\label{tab:single_comparison}
\vspace{-0.5cm}
\end{table}

The distribution shift issue can manifest at any chosen $\alpha$. \ihab{For example, when $\alpha$ is very small, we are likely to switch from $\pi^{\text{nom}}$ to $\pi_{\theta}^\text{safe}$ when the ego vehicle is still very far from the two cars and road boundaries.} During training, the trajectories used to learn $\pi_{\theta}^\text{safe}$ were collected using the fixed switching condition of $V_\theta(f(x, \pi^{\text{nom}}(x))) \leq 0$. This means $\pi_{\theta}^\text{safe}$ was primarily exposed to states where the nominal controller was about to enter the unsafe set. When deployed with a more conservative switching threshold (small $\alpha$), $\pi_{\theta}^\text{safe}$ encounters out-of-distribution states at which it might have 
unpredictable outputs and sometimes cause unsafe behavior. This issue likely affects all member models to varying degrees but is particularly pronounced for member model~2. As shown in Table~\ref{tab:single_comparison}, at $\alpha = 0.06$, member model~2 has the highest $\hat{q}(\alpha)$ of $2.37$. This excessive conservatism causes member model~2 to activate $\pi_{\theta}^\text{safe}$ too frequently, resulting in a 22\% violation rate compared to only 2\% for member model~3. We note that both member models 2 and 3 improve safety compared to only using $\pi^{\text{nom}}$ which achieves a 46\% violation rate. 

These results lead to two key developments in our framework: (1) the ensemble approach discussed next, which aims to improve robustness by aggregating multiple independently trained models, and (2) Phase~2 (safety verification using conformal prediction), which provides 
statistical safety guarantees under the actual deployment distribution.

\subsection{Analysis of the ensemble of HJ value functions used as a safety filter}
A key observation in Figure~\ref{fig:robustness} is that the ensemble does not necessarily outperform its best individual member model, but it consistently achieves very similar performance to the best member model even when the ensemble includes members that perform significantly worse. 

As shown in Table~\ref{tab:single_comparison}, when using all five member models with $\alpha = 0.06$, the ensemble achieves a 96\% success rate and 4\% violation rate, substantially better than the mean member model performance (92\% success, 8\% violation) and matching the performance of member model~3, the best individual model. Notably, the full ensemble (1-5) maintains this 96\% success rate despite including member model~2, which achieves only 78\% success. This demonstrates that the ensemble's aggregation mechanism effectively identifies and leverages the most reliable model at a given state, mitigating the impact of poorly performing members.

To understand how the ensemble size and the quality of individual models affect performance, Table~\ref{tab:ensemble_comparison} presents results for an ensemble with different numbers of member models. When combining member models with different performances (e.g., ensemble 1-2, where member model~1 performs well but member model~2 poorly), the ensemble successfully matches the better member model's performance, demonstrating robustness to weak members. As more member models are added, the ensemble maintains competitive performance close to that of its best member. The same trend is seen across different choices of $\alpha$.

\begin{table}[h]
\centering
\tiny
\begin{tabular}{ccccccccc}
\toprule
\multirow{3}{*}{member models} & \multicolumn{4}{c}{Violation Rate} & \multicolumn{4}{c}{Success Rate} \\
\cmidrule(lr){2-5}\cmidrule(lr){6-9}
& \multirow{2}{*}{Mean} & \multirow{2}{*}{Best} & \multicolumn{2}{c}{Ensemble} & \multirow{2}{*}{Mean} & \multirow{2}{*}{Best} & \multicolumn{2}{c}{Ensemble} \\
\cmidrule(lr){4-5}\cmidrule(lr){8-9}&&&\texttt{multiple} & \texttt{single}&&&\texttt{multiple} & \texttt{single}\\
\midrule
1-2 & 0.15 ± 0.10 & 0.08 & 0.12 & 0.12 & 0.85 ± 0.10 & 0.92 & 0.88 & 0.88 \\
1-3 & 0.11 ± 0.10 & 0.02 & 0.02 & 0.06 & 0.89 ± 0.10 & 0.98 & 0.98 & 0.94 \\
1-4 & 0.09 ± 0.09 & 0.02 & 0.04 & 0.08 & 0.91 ± 0.09 & 0.98 & 0.96 & 0.92 \\
1-5 & 0.08 ± 0.08 & 0.02 & 0.04 & 0.06 & 0.92 ± 0.08 & 0.98 & 0.96 & 0.94 \\
\bottomrule
\end{tabular}
\caption{\small Violation and success rates across 50 trials using $\pi^{\text{nom}}$ and an ensemble of HJ value functions with $\alpha = 0.06$. Rows indicate member model groups (e.g., "1-2" uses $V_{\theta_1}$ and $V_{\theta_2}$). "Mean" shows the average performance across individual member models in each group. "Best" shows the best-performing individual member model in each group. "Ensemble" shows the performance when using all member models.}
\label{tab:ensemble_comparison}
\vspace{-0.5cm}
\end{table}

\subsection{Evaluating the ensemble using \texttt{single} and \texttt{multiple} switching strategies}
\begin{figure}[h]
    \centering
    \subfigure{
    \label{fig:radar}
    \includegraphics[width = .4\linewidth]{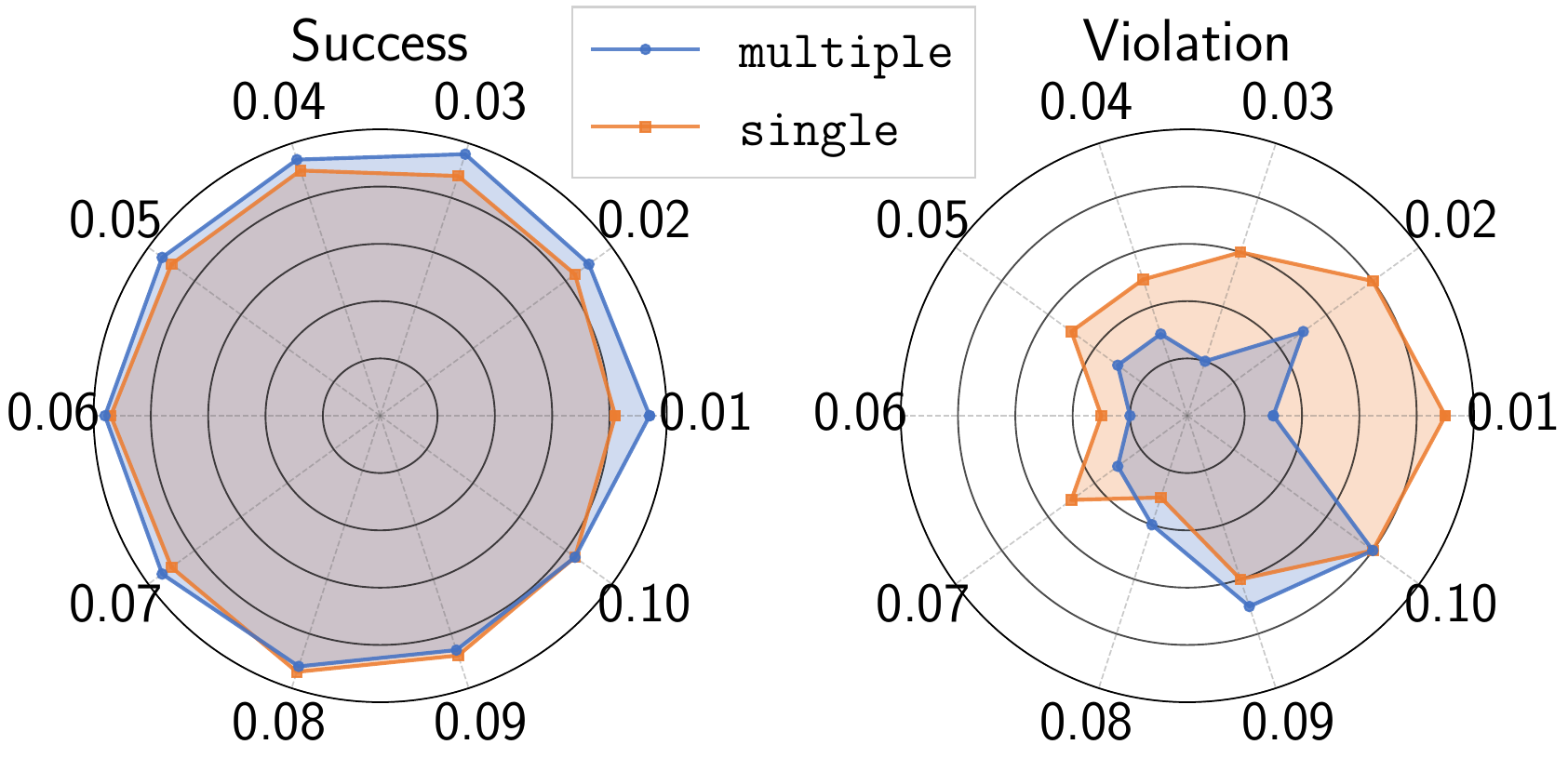}
    }
    \subfigure{
    \label{fig:beta}
    \includegraphics[width = .4\linewidth]{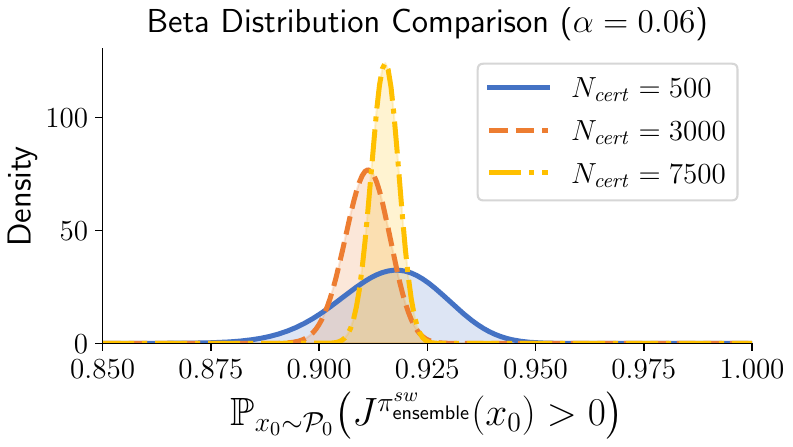}
    }
    \caption{\small(a) The success rate of the safety filters using an ensemble of HJ value functions with different switching strategies (\texttt{single} and \texttt{multiple}) and with different choices of  $\alpha$. The dots in the plot represent the success rates (left figure) and the violation rates (right figure) for the different values of $\alpha$ shown at the perimeter of the circle. The center dot denotes 0, and each black circle represents an increase in value, with a step size of 0.2 for Success and 0.04 for Violation. (b) Graph of $\mathbb{P}_{x_0 \sim \mathcal{P}_0}\big(J^{\pi_\text{ensemble}^{sw}}(x_0) > 0\big) \sim \text{Beta}(N_{\text{cert}} - k, k + 1)$. For each curve, we ran $N_{cert}$ simulations and calculated $k$, the number of simulations where the agent enters the failure set at some time instant in the trajectory, and then plotted the corresponding Beta distribution.}
\vspace{-0.5cm}
\end{figure}

 Comparing the two switching strategies between different ensemble members (\texttt{multiple} and \texttt{single} strategies), Figure~\ref{fig:radar} shows that the \texttt{multiple} strategy generally achieves higher success rates than the \texttt{single} strategy across different values of $\alpha$, with the difference being particularly pronounced at smaller $\alpha$ values.  The superior performance of the \texttt{multiple} strategy can be attributed to its dynamic re-evaluation: by selecting the safest model at each timestep, the agent can exit unsafe states faster than when committing to a single model's safe policy throughout an unsafe episode.



\subsection{Statistical guarantees on the safety of $\pi^\text{sw}_{ensemble}$ and $\pi^\text{sw}$}



In Section~\ref{sec:phase2}, we introduced the conformal prediction safety guarantees for $\pi_\text{ensemble}^{sw}$ and $\pi^\text{sw}$. We evaluate $\pi_\text{ensemble}^{sw}$ using the \texttt{multiple} strategy and $\alpha = 0.06$ and demonstrate that with more evaluations (higher $N_{cert}$), the statistical guarantee becomes more reliable and the Beta distribution concentrates around its mean (Figure~\ref{fig:beta}). This concentration reduces the variance of the safety probability $\mathbb{P}_{x_0 \sim \mathcal{P}_0}\big(J^{\pi_\text{ensemble}^{sw}}(x_0) > 0\big)$, providing more precise safety guarantees.


\section{Related Work}

\label{sec:relatedwork}

The closest work to ours is the very recent work by  ~\cite{somil2025reachability}, where they train neural control barrier-value functions (CBVFs) using physics-informed neural networks and employ conformal prediction to expand their level sets, enabling formulating quadratic programs that are solved online to 
compute the 
closest control to the nominal one that is also safe at each timestep. Our work focuses on HJ value functions rather than CBVFs. 
 \cite{somil2024reachability} uses conformal prediction to verify the superlevel sets of learned backward reachable tubes, focusing exclusively on assuring safety.  
 In contrast, our setting considers both a reward-driven objective optimized for by the nominal controller in addition to safety assurance. 
 Existing works that use HJ value functions as safety filters typically switch to the safe controller whenever $V(f(x, \pi^{\text{nom}}(x))) \leq \epsilon$, where $\epsilon > 0$ is a user-defined threshold without formal guidance on how to choose it~ (\cite{andrea_latentsafety_uncertainty,andrea_latentsafety,tabbara2025designinglatentsafetyfilters}). These methods do not provide formal guarantees on the safety of the resulting policy that arises from switching between the nominal and the unverified learned HJ-based safe controllers. In contrast, we introduced a principled approach that determines the switching condition using conformal prediction, calibrated on the \emph{achievable} safety performance of the learned HJ safe policy, and we derived safety guarantees to certify that the overall switching-based policy keeps the system safe. 
 
\section{Conclusion}
\label{sec:conclusion}
We introduced a two-stage conformal prediction framework that provides probabilistic safety guarantees when using learning-based Hamilton-Jacobi reachability-based safety filters. Our approach addresses the fundamental challenge that learned HJ value functions may be inaccurate by: (1) modifying the switching condition between the nominal and learned safe controllers to account for such inaccuracies using conformal prediction, and (2) verifying the safety of the system under the actual deployment distribution resulting from the developed switching strategy. We also investigated an ensemble-based approach with different switching strategies among the member models. The ensemble demonstrated robustness against low-performing member models by consistently achieving performance close to the best individual models while mitigating the impact of poorly performing members, though it generally did not outperform the best individual policy.

\section{Limitations and Future Work}
\label{sec:limitations}

 An interesting future direction is integrating conformal prediction into the HJ value function training process, which potentially improves the empirical performance of the safety filters in Phase 1.
 Second, HJ value functions are trained using trajectories collected with a fixed switching threshold, but Phase 1 deployment uses adjusted switching thresholds that result in a deployment distribution that is different from the one from which the calibration set is sampled. 
 Future work should explore training with diverse datasets that include 
 trajectories collected using different switching thresholds to improve robustness to the threshold adjustments induced by conformal prediction. Third, while our ensembles are robust to poorly performing  members, they do not outperform their best member models' performances. Training specialized ensemble members optimized for different state-space regions could exploit this robustness to achieve superior overall performance.

\bibliography{references}
\newpage
\appendix

\section{Task Configuration}\label{app:task}
\begin{figure}[h]
    \centering
    \includegraphics[width=0.5\linewidth]{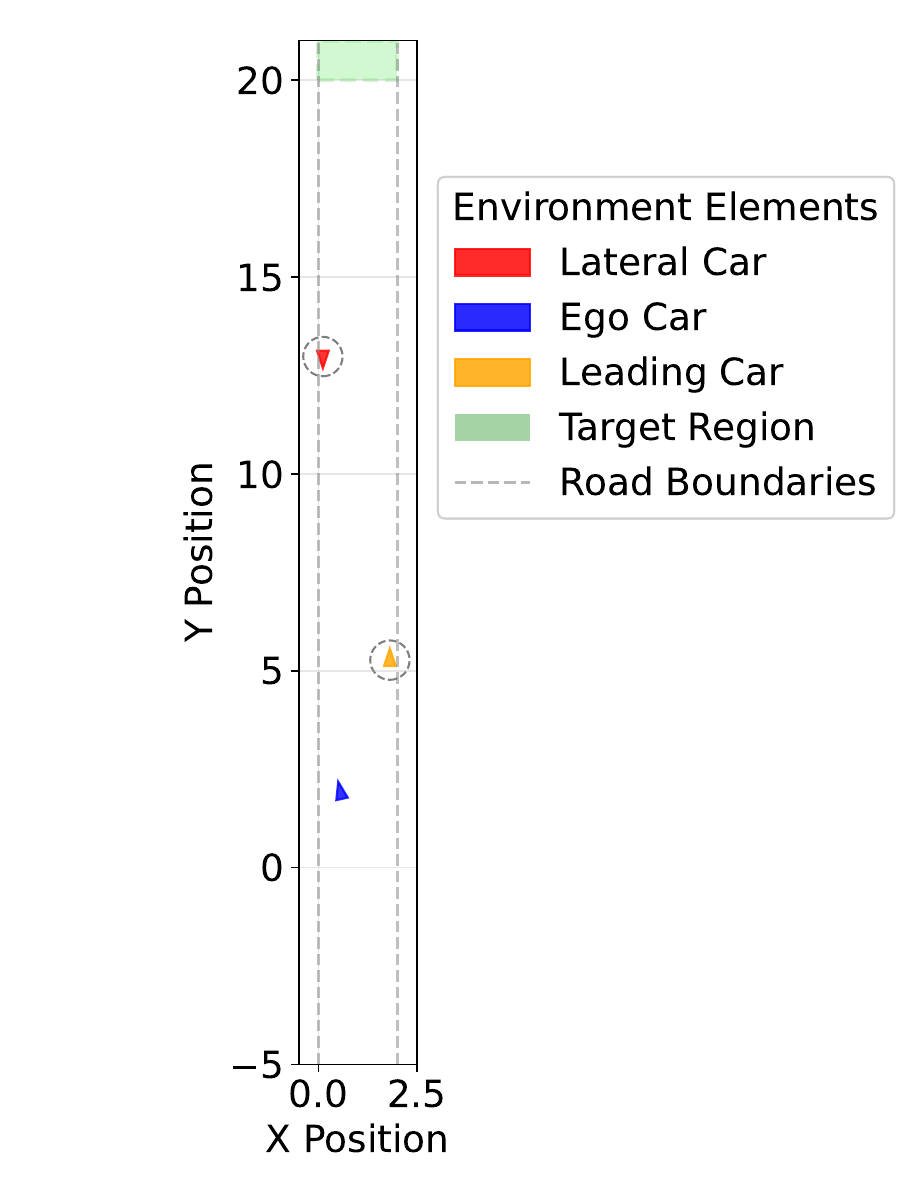}
    \caption{Triple-vehicle Highway Takeover Environment}
    \label{fig:env}
\end{figure}

In the Highway environment, which is shown in Figure~\ref{fig:env}, the state space is defined as
$
    \boldsymbol{\mathrm{x}}=[x_f,y_f,v_f,x_e,y_e,v_e,\theta_e,x_l,y_l,v_l],
$
where $x_f,y_f,v_f$ are the $x,y$ position and the speed of the front vehicle, $x_l,y_l,v_l$ are the $x,y$ position and the speed of the lateral vehicle, and $x_e,y_e,v_e,\theta_e$ the $x,y$ position, the speed, and heading angle of the ego vehicle. The control inputs to the system are the acceleration and angular velocity. 
In each simulation, the vehicles spawn at different locations and move at different speeds. The initial state is randomly sampled with
$
    \boldsymbol{\mathrm{x}} \sim \mathrm{Uniform}\!\left([{\bf a}, {\bf b}]\right),\qquad
$
where ${\bf a}=\bigl(0,\,10,\,0.5,\; 0.3,\,0,\,0.5,\; \tfrac{\pi}{4},\; 1.6,\,5,\,0.5\bigr)$ and ${\bf b}=\bigl(1,\,15,\,2,\; 1.7,\,5,\,3,\; \tfrac{3\pi}{4},\; 2,\,10,\,1.5\bigr)$. $x_f,x_l,v_f,v_l$ are fixed once the state is initialized.
We manually designed $\pi^{\text{nom}}$ to drive the ego vehicle forward by maintaining heading angle $\theta_e \approx \pi/2$, target speed $v_e \approx 2.0$, and $x_e \approx 1$, but without avoiding the two vehicles. A violation is defined as the ego vehicle colliding with another vehicle or leaving the road. A simulation is considered successful when the ego vehicle safely overtakes the two vehicles and reaches the end of the highway (when $y_e\geq20$) within $200$ time steps and without violating any constraints.  Then, we train five HJ value functions $V_{\theta_j}$ and their corresponding $\pi_{\text{safe}}^j$ using DDPG. Each of the five models was trained separately using a different random seed for neural network parameter initialization but with the same training hyperparameters. 
The function $\forall {\bf x} \in X, h: X \rightarrow \mathbb{R}$ used to define the failure set $\mathcal{F} = \{x \mid h(x) \leq 0\}$
is defined as follows: $h({\bf x}) = 10 \min\{\sqrt{(x_e - x_f)^2 + (y_e - y_f)^2} - 0.5,\; \sqrt{(x_e - x_l)^2 + (y_e - y_l)^2} - 0.5,\; x_e - x_{\min},\; x_{\max} - x_e\}$, where $x_{\min} = 0$ and $x_{\max} = 2$ define the road boundaries, and $0.5$ represents the distance between vehicles below which we consider that a collision between them occurred.

\section{Proof of Theorem \ref{thm:conformal_pi_star}}
\label{app:conformal_pi_star}

\begin{proof}
 We set the conformal prediction parameters as follows: the input is $x = x_0$ (initial state), and the score function is $s(x_0, y) =  -J^{\pi}(x_0)$ representing the negative trajectory safety margin under policy $\pi$. The calibration set size is $n = N_{\text{cert}}$, and $\alpha = \frac{k+1}{N_{\text{cert}}+1}$.

The conformal quantile $\hat{q}(\alpha)$ is computed as the $\frac{\lceil(N_{\text{cert}}+1)(1-\alpha)\rceil}{N_{\text{cert}}}$ quantile of the calibration scores $-J^{\pi}(x_0^{(1)}), \ldots, -J^{\pi}(x_0^{(N_{\text{cert}})})$.

This quantile evaluates to:
\begin{align}
\frac{\lceil(N_{\text{cert}}+1)(1-\alpha)\rceil}{N_{\text{cert}}} &= \frac{\lceil(N_{\text{cert}}+1)\left(1-\frac{k+1}{N_{\text{cert}}+1}\right)\rceil}{N_{\text{cert}}} \\
&= \frac{\lceil(N_{\text{cert}}+1)\left(\frac{N_{\text{cert}}-k}{N_{\text{cert}}+1}\right)\rceil}{N_{\text{cert}}} \\
&= \frac{N_{\text{cert}}-k}{N_{\text{cert}}}
\end{align}

Recall that $k$ is defined as the number of calibration scores $-J^{\pi}(x_0^{(i)}) \geq 0$ (equivalently, $J^{\pi}(x_0^{(i)}) \leq 0$, representing that at least 1 state in this trajectory was in the failure set). 
This quantile selects the largest negative calibration score (i.e., the boundary between safe and unsafe trajectories), so $\hat{q}(\alpha) < 0$.

 Following (\cite{angelopoulos2021gentle}), the standard marginal conformal coverage guarantee, not  conditional on any specific calibration set is:
\begin{align}
\mathbb{P}_{(x_0^{(1:N_{\text{cert}})}, x_0) \sim \mathcal{P}_0}\left(-J^{\pi}(x_0) \leq \hat{q}(\alpha)\right) &\geq 1 - \alpha \\
\mathbb{P}_{(x_0^{(1:N_{\text{cert}})}, x_0) \sim \mathcal{P}_0}\left(J^{\pi}(x_0) \geq -\hat{q}(\alpha)\right) &\geq 1 - \frac{k+1}{N_{\text{cert}}+1} \\
\mathbb{P}_{(x_0^{(1:N_{\text{cert}})}, x_0) \sim \mathcal{P}_0}\left(J^{\pi}(x_0) > 0\right) &\geq \frac{N_{\text{cert}}-k}{N_{\text{cert}}+1}
\end{align}

The last inequality follows because if $J^{\pi}(x_0) \geq -\hat{q}(\alpha)$ and $\hat{q}(\alpha) < 0$, then certainly $J^{\pi}(x_0) > 0$.

Conditioning on a fixed calibration set and  following  Section 3.2 in (\cite{angelopoulos2021gentle}), the conditional coverage probability follows a Beta distribution:
\begin{align}
\mathbb{P}_{x_0 \sim \mathcal{P}_0}\left(J^{\pi}(x_0) > 0\right) &\sim \text{Beta}(N_{\text{cert}} + 1 - l, l) \\
\text{where } l &= \lfloor(N_{\text{cert}} + 1)\alpha\rfloor \\
&= \lfloor(N_{\text{cert}} + 1) \cdot \frac{k+1}{N_{\text{cert}}+1}\rfloor \\
&= \lfloor k + 1 \rfloor = k + 1
\end{align}

Therefore:
\[
\mathbb{P}_{x_0 \sim \mathcal{P}_0}\left(J^{\pi}(x_0) > 0\right) \sim \text{Beta}(N_{\text{cert}} - k, k + 1)
\]
\end{proof}

\end{document}